# Detection of Machine-Generated Text: Literature Survey

DMYTRO VALIAIEV, University of Arkansas at Little Rock

Since language models produce fake text quickly and easily, there is an oversupply of such content in the public domain. The degree of sophistication and writing style has reached a point where differentiating between human-authored and machine-generated content is nearly impossible. As a result, works generated by language models rather than human authors have gained significant media attention and stirred controversy. Concerns regarding the possible influence of advanced language models on society have also arisen, needing a fuller knowledge of these processes.

Natural language generation (NLG) and generative pre-trained transformer (GPT) models have revolutionized a variety of sectors: the scope not only permeated throughout journalism and customer service but also reached academia. While there are various good uses, such as speeding up the review and correction process for jobs like narrative development, conversational answers, and code autocompletion, there is still the possibility of misuse. Instances of illicit deployment have occurred in the form of manufactured news, fake product evaluations, spam, and phishing, underscoring the necessity for effective detection techniques to limit the bad implications.

To mitigate the hazardous implications that may arise from the use of these models, preventative measures must be implemented, such as providing human agents with the capacity to distinguish between artificially made and human-composed texts utilizing automated systems and possibly reverse-engineered language models. Furthermore, to ensure a balanced and responsible approach, it is critical to have a full grasp of the socio-technological ramifications of these breakthroughs.

This literature survey aims to compile and synthesize accomplishments and developments in the aforementioned work, while also identifying future prospects. It also gives an overview of machine-generated text trends and explores the larger societal implications. Ultimately, this survey intends to contribute to the development of robust and effective approaches for resolving the issues connected with the usage and detection of machine-generated text by exploring the interplay between the capabilities of language models and their possible implications.

## 1 INTRODUCTION

The use of increased computer power and large datasets has made scalable training of language models possible, revolutionizing the area of natural language processing. This advancement, paired with attention-based transformers, has enabled suggested models to overcome word count limits and write complete texts that are difficult to distinguish from human-authored material [15]. These developments have contributed to a growing trend of more complex language models, which has had both beneficial and harmful consequences.

Language model applications have expanded in recent years, resulting to advancements in chatbots and writing assistants that go beyond simple grammar and punctuation aid. These concepts

Author's address: Dmytro Valiaiev, University of Arkansas at Little Rock.

have been implemented in a variety of industries, including customer service, journalism, and creative writing. Despite the apparent advantages, there are worries about possible misuse, prompting the development of detecting techniques to mitigate the negative repercussions.

The potential issues cover a wide range of aspects of daily life, including spam, phishing, social media overload, and fake news [14]. Furthermore, these models have been shown to be misused in professional and academic writing. Language models have recently proved their ability to generate intriguing headlines, conversations, novels, and poetry [10]. The growth of machine-generated writing has sparked worries about its trustworthiness and authenticity, as well as ethical considerations regarding its use.

One of the most serious possible concerns is astroturfing, in which artificial intelligence-driven social media comment flooding can reflect certain feelings and attitudes [33]. SOTA language models may be enhanced and skewed with extra engineering, allowing them to avoid detection by traditional anti-spam technologies and survive undiscovered in the wild [15]. This trend emphasizes the significance of improving detection tools and increasing public knowledge in order to reduce the impact of malicious use.

Numerous writers have noticed the scarcity of research dedicated to classifying, detecting, and tracking the specific textual signals that might induce people to interpret literature as human-generated rather than machine-generated [2,11,14]. As a result, the general population is still ill-equipped to deal with the inflow of machine-generated text and its consequences. As language models become more widely used, it is critical to focus on the development of detection systems, ethical principles, and public education to guarantee responsible and long-term usage.

## 2 THE INCREASING PREVALENCE OF AI-GENERATED TEXT AND DETECTION CHALLENGES

Rapid advances in artificial intelligence have resulted in tremendous gains in natural language processing. As a result, very complex language models capable of producing human-like prose have been built. This advancement, however, has introduced new hurdles in distinguishing between human-written and machine-generated information, necessitating the development of trustworthy and effective detection systems.

Unlike traditional supervised machine learning models, which compute the probability of y based on x, prompt-based learning focuses on directly producing the probability of a text snippet [6]. Numerous benefits may be realized by eliminating certain sections of the text and iteratively retraining the model with these changed parts. Furthermore, in contrast to the restricted application of labeled factual data used in other models, a large amount of textual data is easily available online [35]. Researchers also investigate an alternative to standard natural language processing, in which the success of a model is dependent on the analyst's feature engineering and domain expertise [30]. This departure involves removing the layer of complexity associated with feature engineering and domain knowledge, allowing the model to learn directly from the massive amount of text available and become analyst-agnostic while still allowing for further refinement [12]. Using fully supervised models, early models in the area concentrated on feature engineering rather than actual predictions, as they were meant to answer concerns like word identification, part-of-speech, and sentence length [9].

The introduction of neural networks accelerated the development of convolutional, recurrent, and self-attentional neural networks [20,25,44].

In 2017, the emphasis turned back to pre-training and fine-tuning, with models charged with forecasting the likelihood of observed data. Tuning then decreases noise and removes meaningful phrases, mostly through masked language modeling and improvements in future sentence predictions [12].

Table 1. Terminology related to thetext generation.

| | |
|---|---|
| Natural Language Generation (NLG) | An artificial intelligence discipline that focuses on creating human-like writing or voice from structured data or other types of input. |
| GPT (Generative Pre-trained Transformer) | A deep learning model that uses transformer architecture to create human-like text in response to a given prompt. GPT models have been iteratively upgraded, with GPT-4 being the most sophisticated version as of 2023. |
| Transformer Architecture | A neural network architecture that is frequently used in natural language processing jobs. Transformers handle input data in parallel using self-attention techniques, which improves efficiency and scalability. |
| Attention Mechanism | A deep learning model approach for weighing the relevance of different bits of input data that allows the model to focus on the most important information throughout the learning process. |
| Fine-tuning | The technique of assigning the weights of a pre-trained neural network to execute a certain job. A big language model, such as GPT, is frequently fine-tuned to fit a specific domain or application. |
| Synthetic Text | Text that is created by artificial intelligence systems, such as language models, rather than by humans. |
| Astroturfing | The process of feigning widespread support or consensus for a concept or cause, sometimes via the use of bogus social media accounts, comments, or reviews. Astroturfing can entail using AI-generated language to produce compelling but fake content. |
| Deepfake | A sort of synthetic material, such as text, photos, or videos, that has been modified or manufactured to seem authentic using artificial intelligence algorithms. AI-generated writing that is difficult to differentiate from human-written information is referred to as deep fake text. |
| Text Classification | A machine learning problem that includes categorizing text based on its content into one or more predetermined classifications. Text classification may be used in the context of AI-generated text detection to assess whether a particular piece of text was created by an AI. |
| Adversarial Attack | A method of manipulating or deceiving machine learning models by providing carefully produced input data that causes the model to produce inaccurate or misleading output. Adversary assaults against AI-generated text might include creating a language that evades identification by text categorization tools. |

By 2021, the pre-train, prompt, predict stage had arisen, in which training tasks were aided by prompts to support the model inefficiently learning from the data [6,35].

To improve model performance, the model attempts to complete incomplete prompts, and suitable prompt engineering is required. The writers distinguish two sorts of prompts: cloze and prefix. Prompts can also be classified as discrete or continuous. The latter, also known as hard prompts, are often natural language words, whereas the former, known as soft prompts, are directly incorporated into the model's space.

Table 2. Explanation of the most popular language models.

| | |
|---|---|
| L2R (Left-to-Right) | L2R language models are generative models that process and create text from left to right. These algorithms forecast the next word in a phrase based on the context of the words that came before it. Many natural language processing applications, such as machine translation and text summarization, have made extensive use of L2R models. |
| Mask | This word relates to a language model approach known as "masking," which is frequently connected with the BERT (Bidirectional Encoder Representations from Transformers) model. Masking entails masking (or concealing) a particular proportion of input tokens at random during the pre-training phase and then asking the model to predict the masked tokens based on the context supplied by the unmasked tokens. This method assists the model in developing a more robust and context-aware grasp of the language. |
| LPM (Language Model Pretraining) | The practice of pretraining a language model on huge volumes of text data before fine-tuning it for specific downstream tasks is referred to as LPM. During pretraining, the model learns to grasp the structure and patterns in the text, such as grammar, syntax, and semantics. This pre-trained model may then be fine-tuned on smaller, task-specific datasets to improve performance in a variety of natural language processing tasks including sentiment analysis, text categorization, and question answering. |
| Encoder-Decoder (Encoder-Decoder) | The Encoder-Decoder architecture, sometimes known as En-De, is a typical technique used in language models and machine translation systems. This design is made up of two main components: an encoder, which analyzes the input text and creates a context vector or hidden representation, and a decoder, which generates the output text based on this hidden representation. The Encoder-Decoder system is very useful for applications like machine translation, where it is used to map input sequences in one language to output sequences in another. |

The writers go through prompt tactics and procedures, as well as answer engineering and answer space design methodologies. Manual design can entail either unrestricted or limited environments, with the latter being related to multiple-choice difficulties. Label decomposition, prune-then-search, and paraphrasing are examples of discrete answer search procedures. Promptless fine-tuning (used in BERT), tuning-free prompting (used in GPT-3), fixed language model prompt tuning, and prompt and language model fine-tuning (used in PADA) are among the training strategies used [6,12].

Knowledge probing techniques such as semantic parsing and linguistic probing typically use pre-train, prompt, and forecast. Their capabilities also include classification jobs such as text categorization and information extraction. The commonsense thinking is a hotly debated issue in the business. Finally, there are many tasks that go beyond classification and generation: the entanglement of template and answer, and the significant work required for prompt answer engineering, especially when dealing with long answers, multiple-class classification problems, or multiple-choice answers.

Table 3. Grouping of pre-trained language models.

| LM | Model Group |
|---|---|
| L2R | GPT |
| | ELMo |
| | XLNet |
| | PanGu-$\alpha$ |
| Mas | BERT |
| | ERNIE |
| | XLM |
| | ELECTRA |
| | BART |
| En-De | T5 |
| | MASS |
| | PEGASUS |
| | GROVER |
| LPM | XLM |

## 3 GENERATED TEXT DETECTION TECHNIQUES: TRANSFORMER ARCHITECTURE AND DECODING STRATEGIES

Modern language models are generally built on the transformer architecture, which excels at dealing with long-term dependencies in text. According to research, unidirectional left-to-right models create more cohesive language [5]. The context is employed in these models to estimate the probability distribution of distinct vocabulary items, and the model then decodes the next phrase in the sequence [1]. Using decoding algorithms that randomly choose tokens from the output distribution frequently results in more general language that is more difficult for people to detect as synthetic [7]. Machine algorithms, on the other hand, can recognize such sentences more readily. The presence of more unusual terms in a text makes it easier for humans to understand.

### 3.1 Text Detection Using Machine Learning

To recognize and detect AI-generated text, several machine learning models and algorithms have been developed. Fine-tuned BERT, Bag-of-Words, Histogram-of-Likelihood Ranks, and Total Probability are among the most famous models [10]. For this objective, logistic regression trained on term frequency-inverse document frequency (tf-idf) characteristics has also been used [23]. Another option is to use Support Vector Machine classifiers, which assess text features such as semantics and grammar. While these classifiers are simple to deploy and need little pre-training, altering them requires extensive re-training [15].

Due to significant developments in natural language generation models such as GPT-3, the identification of artificially created text, as opposed to human-written material, has become an increasingly essential study subject. The capacity to distinguish between human and machine-generated material is critical in a variety of applications, including spam detection, identifying false news, and assuring the legitimacy of online information. To overcome this issue, machine learning approaches have been used to find minor patterns and traits that distinguish produced text from human-written material.

There are possible detrimental effects of false news created by sophisticated language models and provide defense techniques [48]. Hence, the significance of establishing powerful detection methods to safeguard the integrity of online data should not be undermined. The authors advise investing in research to better understand the risks and biases prevalent in cutting-edge language models.

Many researchers studied the effectiveness of existing detection methods in detecting artificially generated language, with a particular emphasis on the text generated by Open AI's GPT-3 [23]. They discovered that current models had variable degrees of effectiveness in recognizing machine-generated text, with some models doing well and others struggling.

One of the methods offered in 2020 for detecting machine-generated text using "BERTscore" [22]. BERTscore is an evaluation measure that uses BERT, a strong language representation model, to quantify the semantic similarity between generated and reference text. The scientists discovered that BERTscore-based classifiers could distinguish between human-written and machine-generated text with high detection accuracy.

## 3.2 Authorship and Fake News Detection Using Stylometry and Feature-Based Classification

As the dissemination of disinformation has become a major worry in today's digital world, authorship and false news detection utilizing stylometry and feature-based categorization have arisen as prominent study fields. The recognition of writing styles and distinctive linguistic elements can aid in the attribution of authorship to anonymous writings and the detection of fraudulent news stories.

In feature-based text categorization, stylometry, a technique that extracts stylistic elements to establish authorship and even identify bogus news, is frequently used [42]. This method has the benefit of being explainable, which other strategies may not have [2]. Classical generators, such as SciGen, created unusual paraphrases that humans can easily spot, whereas discriminator models, like BERT, emphasize statistical abnormalities that are less obvious to unskilled readers [24]. The are several stylometric approaches and machine learning algorithms used in authorship attribution tasks but they are perpetually outrun by the development of newer and more complex models [36]. Indeed, there is the need of establishing unique linguistic traits to distinguish between various writers, as well as the possible application of these approaches in the identification of false news. To identify bogus news items, some authors present a deep learning model that integrates stylometric and content-based properties. They proved that their model surpassed numerous cutting-edge classifiers in terms of detection accuracy [38].

Some authors investigate the use of feature-based classification approaches for detecting troll remarks in online news forums [29]. They construct classifiers that can detect trolls using a range of textual, emotional, and network variables, demonstrating the adaptability of feature-based classification in detecting hostile information online.

Furthermore, some researchers study the application of feature-based classification in identifying and validating rumors on social media platforms [45]. They extract information from tweets' content as well as their social environment and utilize these features to build a machine-learning classifier. Their findings reveal that the classifier can detect and verify rumors, proving the utility of feature-based classification approaches in tackling the problem of fake news on social media.

The relevance of stylometry and feature-based classification algorithms in authorship attribution and false news detection is emphasized by this research. Researchers can design effective methods to minimize the spread of misinformation and maintain the legitimacy of online content by extracting unique language properties and utilizing machine learning techniques.

### 3.3 Human Text Detection Graders and Accuracy Struggles

The increasing intricacy of computer-generated texts has prompted worries about human graders' capacity to recognize such information effectively. Recent research has looked at this topic, examining the ability of human assessors to distinguish between human-written and machine-generated writings.

Human graders have been used in several research to assess and confirm the outcomes of machine-generated text detection. Amazon Mechanical Turk and crowdsourcing through websites like Real or Fake Text (RoFT) have proven popular ways to recruit human assessors [10]. These investigations, however, have revealed that even skilled human raters struggle to discern synthetic texts as reliably as computer discriminators [23]. Language models are predicted to get more complex as computing power develops, thereby lowering the accuracy of human graders. Some authors assess human graders' abilities to detect machine-generated texts [13]. Participants struggled to tell the difference between human-written and neural language model-generated texts, underscoring the potential hurdles offered by increasingly improved artificial language creation. Certain researchers investigate the effectiveness of a statistical technique in aiding human graders in recognizing artificially created texts [16]. The authors present a visual tool that indicates the possibility of a language model producing each word. Despite the fact that the tool enhances detection accuracy, the study underscores the ongoing issue of distinguishing between human and machine-generated texts. Additionally, there are studies that explore the performance of human assessors in detecting texts created by cutting-edge language models [43]. The study found that evaluators failed to discern between human-written and machine-produced writings, adding to worries about the increasing difficulty of spotting artificially generated information. While human performance was restricted, they discovered that automatic systems may reach greater detection accuracy, indicating a possible path for enhancing detecting skills [22].

These studies demonstrate the difficulties that human graders experience when spotting computer-created writings. As language models improve, it is critical to provide new tools and approaches to help human assessors and enhance the whole detection process, preserving the quality of online information.

### 3.4 AI-generated Text Assessment and Editing Tools and Techniques: Combining Generation and Detection Capabilities

Because of the rapid improvements in AI-generated text, there has been a surge in interest in developing evaluation and editing tools that can not only detect but also refine machine-generated material. Several types of research have been conducted to investigate the combination of generation and detection skills in order to improve the quality and reliability of AI-generated text. For instance, the Giant Language Model Test Room (GLTR) is a solution that takes into account word likelihood, rank, and entropy [3]. Some language models, such as GROVER, combine production and detection capabilities, although their detection of the text created by other models is restricted [3]. Tools such as the KNOWLEDGEEDITOR have been proposed to reduce time and effort on retraining or fine-tuning by enabling users to modify particular facts in the produced text, raising ethical issues about inserting inaccurate information into language models [4]. Moreover, some researchers are exploring deep learning strategies for detecting style, sentiment, and text coherence [1]. It is noteworthy to highlight these models have the potential to overall increase the quality of AI-generated content by detecting and resolving inconsistencies, errors, and aesthetic flaws. Additionally, managing text production by including prompt-specific information can significantly improve the quality of the generated material [8]. In this case, a generative pre-trained transformer model may be guided by particular keywords and features to ensure the generated text has the appropriate qualities and quality. It is also popular to use a two-step technique for producing and revising content using language models [11]. This technique seeks to increase the overall quality and coherence of the AI-generated content by first creating a draft and then repeatedly revising it, opening the way for more effective editing tools. This is a somewhat groundbreaking approach as combining detection and correction approaches in a single framework preemptively solves the problem of authorship attribution.

## 4   POTENTIAL APPLICATIONS AND FUTURE DIRECTIONS

As AI-generated text becomes more common, it is critical to create and improve detection systems in order to reduce the hazards associated with the dissemination of misleading and harmful material. Researchers may investigate novel approaches, such as infusing human intuition and knowledge into machine learning algorithms or constructing hybrid models that combine the capabilities of several methodologies, in addition to improving existing detection models. Recent research has extensively examined the possible uses of AI-generated text detection and prospects in the area, with several papers emphasizing rising patterns and new possibilities. Many authors analyze the application of deep learning techniques in detecting and preventing the spread of false news [31]. The report addresses the fundamental obstacles and future possibilities for artificial intelligence-generated text identification in the context of misinformation and disinformation. Given the rising sophistication of generation models and their possible misuse, some authors emphasize the necessity for more study in recognizing AI-generated material [18]. This is an example of another research that presents a summary of the current state of the art in text creation techniques and their possible applications. As was mentioned before, the investigation of the function of AI-generated text identification in the context of automated fact-checking is seen as the most proliferate next stage in developing the algorithms [19]. These researchers address the field's difficulties and potential, emphasizing the importance of collaborating to create more effective detection systems to help fact-check programs. Additionally, there is extensive research that is done towards the examination of the possible benefits and limitations of incorporating AI-generated text detection into human-AI collaborative writing systems [26]. The authors present a framework for merging AI-generated text evaluation and editing tools, highlighting possible applications and future research in this field.

   These studies show the broad range of possible applications for AI-generated text identification and the need for continued study in this field. As AI-generated text becomes more common and powerful, the development of strong and effective detection systems will become increasingly important in resolving ethical problems, combatting disinformation, and enabling human-AI collaboration across several domains. Furthermore, when AI-generated text identification improves, its uses may expand beyond disinformation detection. It might, for example, be used to analyze the quality of machine-generated material in industries such as journalism, marketing, and entertainment, guaranteeing that AI-generated content satisfies human-generated work standards.

## 5   ETHICAL ISSUES WITH AI TEXT DETECTION LIMITATIONS: MISINFORMATION AND MALICIOUS CONTENT

Researchers and practitioners are increasingly concerned about the ethical implications of AI text detection limits, particularly in the context of misleading and harmful content. Several pieces of research have addressed these issues, highlighting the importance of strong detection methods as well as an ethical framework for AI applications.

   Ethical considerations develop when AI-generated text grows more complex. The possibility of generating and disseminating misleading information or harmful content is becoming more prevalent. Furthermore, the availability of pre-trained language models expressly calibrated for detection remains restricted, as these models need considerable pre-training and might be challenging to deploy across many architectures [3]. Nowadays one of the most popular approaches is to examine information diffusion on social media platforms and emphasize the essential role of AI in limiting the transmission of misleading information [46]. This particular study underlines AI developers' ethical obligation to construct trustworthy and accurate detection systems in order to prevent misinformation from spreading. Some authors investigate the possible misuse of AI in different sectors, including the production of false text [7]. The study underlines the significance of multidisciplinary collaboration in understanding and addressing AI's ethical concerns, and it recommends best practices for AI developers and policymakers. It also should be noted that while models like BERT have achieved outstanding outcomes in numerous NLP tasks, they are nevertheless prone to mistakes and restrictions [41]. These constraints may result in AI-generated disinformation or difficulties recognizing dangerous material, highlighting the need for more study and development to address the ethical considerations related to AI text detection. To provide insights into machine learning fairness and ethical issues, many researchers are convinced that the need of ensuring AI models are impartial and transparent should be stressed [4]. This viewpoint is especially pertinent in the context of AI text recognition, where ethical issues regarding the possible amplification of prejudice and disinformation must be addressed.

Hence, it can be concluded that numerous studies highlight the ethical concerns related to AI text identification limits, as well as the possibility of disinformation and harmful content. As AI technology progresses, the creation of strong and transparent text detection algorithms will be critical in resolving these ethical issues and assuring the appropriate use of AI in many applications.

## 6 TIMELINE OF RESEARCH AND TRENDS

Since the introduction of GPT-2 in 2019, in particular, interest in identifying and masking fake text has risen significantly

### 6.1 Before 2019

In 2018, the biggest trend was associated with evaluating the performance of language models. The creation and use of extensive pre-trained models as well as the introduction of the transformer architecture dominated the research trend in language models. Enhancing language comprehension and natural language creation ability was the main goal.

The creation and use of extensive pre-trained models and the introduction of transformer architecture were the main focuses of language model research in 2018 [44]. The major emphasis was on strengthening language comprehension and natural language-generating skills. Vaswani et al. presented the transformer architecture, which by applying a revolutionary self-attention mechanism greatly enhanced the performance of several natural language processing tasks. This pioneering study served as the basis for several language models created in 2018 and beyond. BERT (Bidirectional Encoder Representations from Transformers), a groundbreaking paradigm that drew on the transformer design, was suggested by Devlin et al. [12]. BERT obtained cutting-edge performance on a variety of NLP tasks by utilizing unsupervised pre-training on a sizable corpus of text.An early version of the well-liked GPT-2 and GPT-3 models, GPT (Generative Pre-trained Transformer) was introduced by Radford et al. [35]. To obtain good performance on a variety of NLP tasks, GPT used unsupervised pre-training and fine-tuning. ELMo (Embeddings from Language Models), a model that gave deeply contextualized word representations and allowed for a better comprehension of polysemy and context-dependent meanings, was introduced by Peters et al. [34]. A significant step forward in the creation of pre-trained language models, ELMo showed significant gains on a range of NLP tasks.

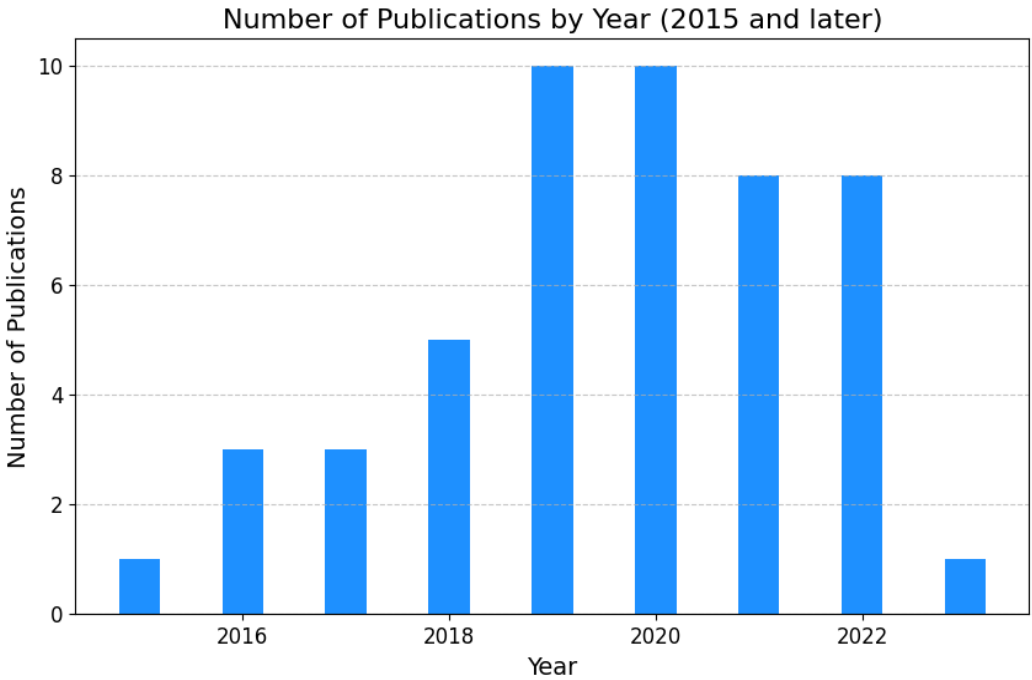

Fig. 1. 1. Number of publications used in this survey by year.

Large-scale pre-trained models and transformer-based architectures remained the focus of language model research in 2019. The emphasis shifted even more to optimizing these models for diverse NLP tasks, correcting their shortcomings, and investigating their possible uses. The following peer-reviewed journal papers and conference proceedings provide examples of this pattern: GPT-2, an upgraded version of GPT that performed well on a variety of NLP tasks [35]. The scalability and efficacy of GPT-2 generated debates on the advantages and disadvantages of large-scale language models as well as their moral ramifications. In 2019, Liu et al. developed RoBERTa, an improved version of BERT that produced cutting-edge results on a number of benchmark tasks [27]. To surpass the original BERT model, RoBERTa added additional changes including longer training and bigger batch sizes. To address the shortcomings of BERT, Yang et al. unveiled XLNet, a generalized autoregressive pre-training model [47]. XLNet outperformed BERT on a number of NLP benchmarks by including permutation-based language modeling.

### 6.2 2020

Major trends:
- Automatic text generation detection.
- The effects of language models on society.
- Discriminatory text produced by a machine.

The development of large-scale pre-trained models and transformer-based architectures continues to serve as a foundation for the research trend in recognizing autonomously created texts in 2020. To better identify artificially created information, eliminate any biases, and increase their comprehension of the inner workings of the models, researchers started to investigate a variety of strategies. The following peer-reviewed journal papers and conference proceedings provide examples of this pattern:

To identify machine-generated material, some researchers started looking into the issue of fine-tuning big language models like GPT-2 [40]. The use of supplementary tasks for training text generation models to strengthen them against hostile cases was investigated by Zhang et al. [50]. The authors showed that the ensemble model could recognize bogus text better than individual models.

### 6.3  2021

Major trends:
- Statistical text generation detection and visualization.
- Examples of oppositional arguments for assessing NLU systems.
- Separating artificially generated text from real language.

Research on identifying texts produced by neural networks in 2021 focused on improving detection methods, comprehending the drawbacks of current approaches, and resolving potential biases in massively pre-trained models. Studies investigated numerous methods to improve the effectiveness of text detection systems, as shown by several notable papers. For displaying and identifying the outputs of massive language models, Gehrmann et al. introduced a technique dubbed GLTR (Giant Language Model Test Room) [16]. The authors showed that GLTR could successfully identify between text produced by humans and text produced by machines using probability-based ranking.

## 7  DISCIPLINES AND COLLABORATION

### 7.1  Disciplines

Interdisciplinary approaches to artificial text identification have also aided study, with experts from linguistics, computer science, and psychology sharing their knowledge. For example, some authors investigated the relevance of linguistic signals in recognizing AI-generated writing [28]. In this work, some researchers suggest data statements as a design solution and professional practice for natural language processing technologists in research and development.

The discipline may begin to address key scientific and ethical challenges that arise from the use of data from particular populations in the creation of technologies for other communities by the adoption and broad usage of data declarations [3]. Computer science, linguistics, information security, and data privacy all contribute to the multidisciplinary nature of this research. The bulk of articles come in computer science journals and conferences; however, linguistics and information security venues are seeing an increase in the number of publications.

Researchers in computer science have mostly concentrated on building algorithms and models for recognition. For example, some studies give a comprehensive assessment of the deep learning approaches employed for this purpose [21]. Similarly, others give a tool for recognizing produced text by assessing the likelihood of word selections in a given text [16].

Linguistics is also important in spotting falsely made text. The study of language style, and stylometry, has been used to distinguish between human-authored and machine-generated writing. Some older research examines the historical backdrop as well as several stylometric approaches for attribution, which can be expanded to identify the AI-generated writing [36].

The societal ramifications of AI-generated writing and its detection have been studied in the social sciences and communication studies. Some studies investigate the function of AI-generated text identification within the framework of automated fact-checking, emphasizing its significance in countering misinformation and deception [19]. Furthermore, others examine the present state of AI-generated text detection and consider the consequences for various stakeholders, such as journalists, legislators, and educators [24].

Finally, the discipline of ethics has addressed the moral issues and consequences of AI-generated text identification. Some researchers explore the ethical quandaries that come from the usage and possible misuse of AI-generated text, highlighting the importance of responsible detection technique development and deployment [5].

Overall, the disciplinary distribution of research on identifying artificially created text exemplifies the field's multidisciplinary nature, with contributions from computer science, linguistics, social sciences, communication studies, and ethics, among others. This range of viewpoints emphasizes the complexities and significance of tackling the issues and opportunities related to AI-generated text recognition.

### 7.2 Collaboration

join togethertransdisciplinary and significant research.

Fig. 2: Co-authorship network of the authors in this survey.

These examples show how research on identifying artificially created texts is collaborative in nature. Researchers from a variety of fields, including linguistics, computer science, natural language processing, and data science, frequently collaborate to create cutting-edge methods and strategies for overcoming the difficulties presented by automated text production. Some authors showcase cooperation between academic and research institutions by bringing together scholars from the University of Washington and the Allen Institute for Artificial Intelligence [48]. Other researchers illustrate how academic institutions and industrial research laboratories may collaborate [17]. Zhang et al., who created a deep learning-based text classification model, were among the early researchers to examine strategies for identifying artificially generated text [49]. Their study established the groundwork for future research into more complex ways to distinguish between human-generated and AI-generated text.

## 8   CONCLUSIONS

The machine-generated text has the potential to improve many elements of human interaction with textual information. Grammar correction, machine translation, writing support for both textual and programming material, and even poetry production are all examples of useful uses. However, when dealing with bad content, such as misinformation, unwanted messages, and misleading methods, a slew of issues develop. As a result, the capacity to distinguish between human-authored and machine-generated content is critical.

In general, simpler and more readily understood material is less likely to be seen as synthetic. The key dangers are authorship loss, authenticity, and general divergence from the truth. The industry need a simple, feature-based classifier that can distinguish between human and produced content. Grover, for example, can not only produce but also identify generated text. RoBERTa outperforms GPT-2 in detecting tasks, especially with short text snippets like as tweets, but problems with spelling mistakes and homoglyphs.

The development of methods that examine linguistic traits, statistical traits, and the substance of the generated text has been the main goal of research into the detection of fake text.

Popular techniques for detection include:

• Feature-based strategies These methods use linguistic analysis to distinguish between writing that was produced by humans and text that was generated artificially.

• Statistical techniques In these methods, deviations from natural language are found by analyzing statistical aspects of the text, such as word embeddings or n-gram frequencies.

Co-authorship Network (Last Names Only)

Fig. 2.

• Methods based on machine learning These techniques use machine learning algorithms, either supervised or unsupervised, to categorize synthetic text based on properties retrieved from the text.

The essence of the problem is that current detectors beat supervised spam classifiers, but they struggle to keep up with natural language model developments. Because of lower processing costs and the availability of free data, these models have flourished. As a result, future research should concentrate on constructing customized sub-classifiers for each dataset and use an ensemble technique for output. Many existing interim solutions rely on generator knowledge, which provides difficulty.

False positive findings are especially troubling because they may lead to prejudice against actual human perspectives. Furthermore, detection approaches are sometimes difficult to transfer across several generators. As a result, no final answer exists. Detectors can be temporarily fixed, but they will require ongoing improvement and fine-tuning. More complex models need more resources, making them less accessible to the general population. When looking for an acceptable solution, it is critical to consider variables such as transferability, set up, and convenience of use.